\documentclass{spie}

\usepackage{cite}
\usepackage{times} 
\usepackage{indentfirst} 

\usepackage[colorinlistoftodos,color=pink]{todonotes} 
\usepackage{xkeyval}
\presetkeys{todonotes}{inline}{}

\usepackage{subfigure}
\usepackage{amsmath}
\usepackage{enumerate}
\usepackage{diagbox}
\usepackage{amsfonts}
\usepackage{import}
\usepackage{graphicx}
\usepackage{multirow}
\usepackage{caption}
\usepackage{comment}
\usepackage{array}
\usepackage{flafter}
\usepackage{float}
\usepackage{tikz}
\usetikzlibrary{decorations.pathreplacing,calligraphy}
\usepackage{color}  
\usepackage{hyperref}
\hypersetup{
    colorlinks=true, 
    linktoc=all,     
    linkcolor=red,  
    citecolor=red,
}
\title{HoughToRadon Transform: New Neural Network Layer \\ for Features Improvement in Projection Space}

\author{Alexandra Zhabitskaya\supit{1, 2}, Alexander Sheshkus\supit{2, 3}, and Vladimir L. Arlazarov\supit{2, 3}
	\skiplinehalf
	\normalsize 
	\supit{1} Lomonosov Moscow State University, Moscow, Russia; \\
        \supit{2} Smart Engines Service LLC, Moscow, Russia; \\
        \supit{3} Federal Research Center “Computer Science and Control” of RAS, Moscow, Russia.
}

\begin{document}
	
	\maketitle
	
	\begin{abstract}
		In this paper, we introduce HoughToRadon Transform layer, a novel layer designed to improve the speed of neural networks incorporated with Hough Transform to solve semantic image segmentation problems. By placing it after a Hough Transform layer, 'inner'\ convolutions receive modified feature maps with new beneficial properties, such as a smaller area of processed images and parameter space linearity by angle and shift. These properties were not presented in Hough Transform alone. Furthermore, HoughToRadon Transform layer allows us to adjust the size of intermediate feature maps using two new parameters, thus allowing us to balance the speed and quality of the resulting neural network. Our experiments on the open MIDV-500 dataset show that this new approach leads to time savings in document segmentation tasks and achieves state-of-the-art 97.7\% accuracy, outperforming HoughEncoder with larger computational complexity.
		
	\end{abstract}
	
	\section{Introduction}
	
In the modern world, there is sustained interest in image processing and analysis. Pattern recognition is a priority direction in the development of artificial intelligence. Convolutional neural networks (CNNs) are one of the various methods used to solve this task. Currently, they are being involved in the classification of medical images \cite{lit1}, \cite{lit2}, road signs \cite{lit3}, \cite{lit4} or handwritten numbers \cite{lit5}. A method using SE-ResNet modules was proposed in \cite{lit2} to classify breast cancer tissues. More and more convolutional neural networks are being developed for 3D object detection, as seen in \cite{lit6}. 

Convolutional layers have always been prized for their ability to process local features, but recently there has been a contrary view. In \cite{lit22}, the authors claim that convolutional characteristics, once considered strengths, are now seen as limitations. There are three main issues. Firstly, convolutions process all image pixels regardless of their importance and position. This leads to spatial inefficiency, particularly in image segmentation tasks where certain image objects are prioritized over others. Secondly, high-level features may not always be present in an image, making the use of pre-trained feature filters inefficient. Finally, convolutions struggle to establish dependencies between distant pixels. Each convolutional filter is confined to operating within a small region, but long-range interactions between semantic concepts are crucial in some tasks. To handle spatially-distant concepts, existing approaches increase kernel size or model depth. However, this compensates for the weaknesses of convolutions by adding complexity to the model, which impacts on training time and computational resources. These problems encourage the use of neural networks that rely on tools operating with global features rather than exclusively on convolutional ones. Moreover, most new architectures are designed according to well-known models and are combinations of already studied layers. Searching and discovering new combinations is highly important in addressing a broader range of computer vision tasks.
	
	\section{Hough Transform}
 Hough Transform (HT) is one of the widely used tools for image analysis \cite{lit9}. In the ($x, y$) coordinate space, a line is defined by the slope $\alpha$ and the shift $s$ along the ordinate axis: $y = tg(\alpha) x + s$. Let's consider a mapping between ($x, y$) and ($s, \alpha$) spaces: a straight line in ($x, y$) space is mapped to a point with coordinates ($s, \alpha$). In more detail, if points form any line in the original space, the intersection of Hough lines (points image) provides the desired ($s, \alpha$) values. Each point of the Hough space ($s, \alpha$) is an integral of the pixel intensity along the direction corresponding to the angle $\alpha$ and the shift s. The Hough Transform can be defined using the following formula:

 \newpage

\begin{equation}\label{eq1}
H(s, \alpha) = \sum_{(x, y) \in l(s, \alpha)} I(x, y), 
\end{equation}

where $l(s, \alpha)$ is a line in the original image with angle $\alpha$ and shift s.

Over time, many modifications to the original Hough Transform have been created, significantly changing the appearance and structure of the output image. In \cite{lit11}, the authors posed the following problem: originally, HT was only applied to a fragment of the image, but when analyzing real data, this approach proved to be inconvenient. To span the entire image area with lines, they proposed expanding the original image by $h \times h$ or $w \times w$ (where h and w represent the height and width of the input image). 

The classical Hough Transform has a complexity of O($n^3$), which limits its applicability to large datasets. Fast Hough Transform (FHT) offers a solution here. Operating with dyadic patterns, FHT reduces the complexity to linear-logarithmic O($n^{2}logn$), where n is the image size. The idea is to split the input image into two halves using a bit-shift operation and then apply FHT to each half individually. This method efficiently utilizes the input data's periodicity and symmetry to enhance the computation speed, but requires to divide the range of angles into 4 parts. Obviously, performing FHT for only one angular range (one quadrant) is insufficient for a full analysis of the image. The paper \cite{lit12} suggests a method of combining image quadrants based on their common edge regions. The output image of such a transformation for all 4 quadrants will have a height of $4 \times h - 3$ (when combining images, we subtract a common line, so that the images are "glued"\ together). It turns out that for complete data analysis, it is convenient to use a version of HT that receives an image of size $h \times h$ and produces an image of size $(h + h) \times (4 \times h - 3)$ (for square images of the size of a degree of two). As a result, the area of the feature maps expands by approximately a factor of $8$, leading to a considerable rise in the cost of computing convolution layers on them. The neural network architecture in \cite{lit15} is particularly affected by this problem since the "inner"\ convolutions between FHT and TFHT layers work with enlarged feature maps.

Using Hough Transform as an inner layer for intermediate feature maps is a new trend in developing neural network architectures. For instance, it is applied before training hierarchical neural networks for character recognition \cite{lit19}. Based on Hough Transform, a HoughNet architecture is created to detect vanishing points outside the image \cite{lit7}. Hough Transform played a key role in the development of a neural network for human eye recognition \cite{lit18}. Furthermore, the algorithm is successfully used for semantic document segmentation \cite{lit10}.

In architectures that incorporate a Hough transform layer, post-Hough convolutions are required to extract complex non-linear features along various straight lines. HT converts the image into a space with new coordinates, hence modifying its size. However, assuming a significant increase in the area of the original image, performing post-HT convolutions becomes computationally expensive. Therefore, the issue of image size is quite acute.

\begin{center}
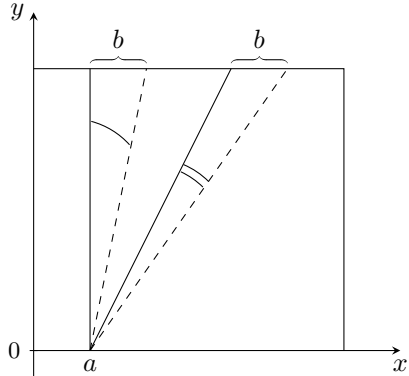
\begin{figure*}[h]
\centering
\begin{tikzpicture}[>=stealth, scale=0.75]
\draw[->, thin] (0,0) -- (6.5,0) node[below] {$x$};
\draw[->, thin] (0,-0.5) -- (0,6) node[left] {$y$};
\draw (0,5) -- (5.5,5) -- (5.5,0);
\draw[dashed] (1, 0) node[below] {$a$} -- (2, 5);
\draw (1.0, 0) -- (3.5, 5);
\draw[dashed] (1.0, 0) -- (4.5, 5);
\draw (1, 0) -- (1, 5);
\draw[black] (1.7,3.65) arc (42:75:1.4);
\draw[black] (3.1,3) arc (45:67:1.4);
\draw[black] (3, 2.9) arc (45.5:65:1.4);
\draw [decorate,
    decoration = {calligraphic brace}] (1,5.1) --  (2,5.1) node[pos=0.5,above=2pt,black]{$b$};
\draw [decorate,
    decoration = {calligraphic brace}] (3.5,5.1) --  (4.5,5.1) node[pos=0.5,above=2pt,black]{$b$};
\foreach \y in {0}
\draw[shift={(0,\y)}] (2pt,0pt) -- (-2pt,0pt) node[left]
{\footnotesize $\y$};
\end{tikzpicture}
\caption{Two pairs of lines with the same (a, b) change correspond to different angles.}
\label{ris:image2}
\end{figure*}
\end{center}

Let us consider another factor that reveals the imperfections of Hough Transform in convolutional networks. In $R^2$ space, any straight line can be uniquely determined by two parameters. The first type of parameterization is as follows: a straight line is specified by the angle $\varphi$ of its normal ($cos\varphi, sin\varphi$) slope to the x-axis (in contrast to $\alpha$ which is an angle between the line itself and the x-axis) and the distance from the origin $\rho$. The ($\rho, \varphi$) plane is also referred to as the Hough space for a set of lines. The second type of parameterization defines a straight line using the coordinates $(s,t)$. For mostly horizontal lines ($45 \leq \varphi < 135$), these parameters specify the y-coordinate of the intersection of the line with the left boundary of the image and its variation from the left to the right boundary, respectively; for mostly vertical lines ($-45 \leq \varphi < 45$), they specify the x-coordinate of the intersection of the line with the bottom boundary of the image and its variation from the top to the bottom boundary, respectively. 

Let the input image be of size $w_{1} \times w_{1}$. Then the following relation \cite{lit16} between the parameter spaces $(s, t)$ and $(\rho, \varphi)$ exists:

\begin{equation}\label{eq2}
tg (p \cdot \varphi) = -(\frac{w_{1}}{t})^{p},  \quad \rho = \frac{s \cdot w_{1}}{\sqrt{t^{2} + w_{1}^{2}}},
\end{equation}

where $p = 1$ for mostly horizontal lines and $p = -1$ for mostly vertical lines. Note that the transition from one coordinate system to another is non-linear in terms of both angle and shift. The linear change in input image coordinates does not correspond to linear changes in angle and shift, which is bad for convolutions that rely on the size of the feature elements in feature maps \cite{lit21}. The non-linearity of the coordinate change is clearly shown in Fig. \ref{ris:image2}.

In this paper, we propose a layer that performs a conversion from $(s, t)$ to $(\rho, \varphi)$ space. We will call it a HoughToRadon Transform layer. This layer addresses two previously mentioned issues, namely the high computational cost of convolution layers after HT and the non-linearity of image coordinates concerning angle and shift, while providing no compromise to the performance of the NN.

The remainder of this paper is structured as follows. Section \ref{sec3} presents a detailed description of the transformation method. Section \ref{sec4} describes the experimental setup, datasets and evaluation metrics used to access the performance of the newly introduced HoughToRadon transform layer. Section \ref{sec6} provides the results of the conducted experiments and the analysis of them. Finally, section \ref{sec7} concludes the paper and outlines potential directions for the future research.

 \section{HoughToRadon Transform}\label{sec3}

Let us describe the principle of converting an image from ($s, t$) space to ($\rho, \varphi$) space and back. For the sake of simplicity, we only consider square images with a side of a degree of $2$, although the following may be extended to images of an arbitrary size with some modifications.

Next, we detail the output image layout, $w_{2}$ and $h_{2}$ represent its width and height respectively. $w_{1} \times w_{1}$ is the size of the FHT input image. Let $n$ denote the number of angles -- the input parameter of HoughToRadon transform. As mentioned above, the image produced by FHT consists of $4$ regions: mostly vertical lines with a slope to the right ($-45 \leq \varphi < 0$), mostly vertical lines with a slope to the left ($0 \leq \varphi < 45$), mostly horizontal lines with a slope downwards ($45 \leq \varphi < 90$) and mostly horizontal lines with a slope upwards ($90 \leq \varphi < 135$). We iterate over the range of angles $[-45; 135]$ with a step of $180/n$, creating a list of angles denoted by $A$. Each angle from the created list, arranged in ascending order, corresponds to one horizontal line in the output image, i.e. $h_{2}$ = $n$. $w_{2}$ is assumed to be equal to the maximum integer radius in the source image before applying FHT, i.e. $\sqrt{2w_{1}^{2}}$.

We perform the transformation by iterating over the output image coordinates to ensure there are no missing cells. An image pixel is defined as ($i, j$), where $i$ ranges from 0 to $w_{2}$ and $j$ ranges from 0 to $n$, with both $i$ and $j$ being positive integers. Let $\rho = i$ and $\varphi$ be the $j$-th element in the ordered list A. We consider all such ($\rho, \varphi$) pairs. For each of them, we have to find a corresponding ($s, t$) pair and retrieve the value of the cell ($s, t$) in the input image.

The image size can be adjusted not only by changing the $n$ value. We also introduce the $scaleX$ parameter and describe how it controls the output image width. Iterating through the coordinates ($i, j$) of the output image, which is now of size $(w_{2} \times scaleX) \times h_{2}$, each cell is filled with the value from the input image cell ($s, t$) for the pair ($i/scaleX, j$). Since the pixel values are discrete, we must round acquired values to the nearest integers ($s,t$). We kept this part simple, assuming that convolution layers learn through local window filters themselves. Using this approach, the image can be conveniently scaled to adjust its size, providing further reduction in model's computation and memory requirements.

Considering the equations \eqref{eq2} and the introduced $scaleX$ parameter, ($\rho, \varphi) \rightarrow (s, t$) mapping is a straightforward task:

\begin{equation}\label{eq4}
    t = - w_{1} \cdot \tan (- \varphi), \quad s = \frac{\rho}{scaleX \cdot w_{1}} \cdot \sqrt{t^{2} + w_{1}^{2}},
\end{equation}

where $-45 \leq \varphi < 45$.

\begin{equation}\label{eq5}
    t = - w_{1} / \tan (\varphi), \quad s = \frac{\rho}{scaleX \cdot w_{1}} \cdot \sqrt{t^{2} + w_{1}^{2}},
\end{equation}

where $45 \leq \varphi < 135$. Equations are given without considering the shift for each of the 4 regions.

We introduce HoughToRadon transform layer (HRT) that performs the described transformation. Since it has no trainable parameters, it does not add to the number of weights of the NN. All operations are predefined and remain constant during training. The transposed layer (RadonToHough transform, RHT) is applied to convert from ($\rho, \varphi$) space back to ($s, t$) space and is also utilized for gradient propagation during the NN training.

\begin{figure}[h]
\centering
\scalebox{0.85}{\subfigure{
\includegraphics[width=\textwidth]{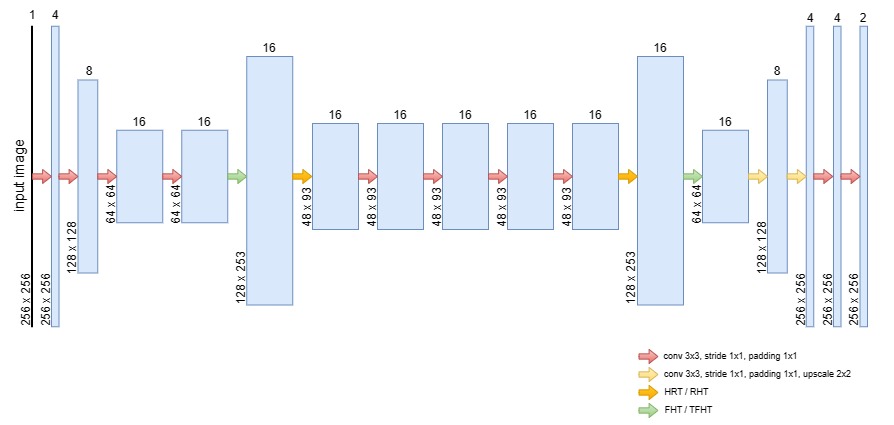}
}}
\caption{Visual interpretation of the new NN.}
\label{ris:nn}
\end{figure}

\section{Experiments}\label{sec4}

\subsection{Implementing in the HoughEncoder architecture}

In this section, we demonstrate the effectiveness and efficiency of the proposed layer by implementing it in the pre-existing HoughEncoder architecture \cite{lit10}. This neural network is used as an autoencoder for semantic segmentation of images with documents, producing an image in the same coordinate space. The additional layers must not translate the Hough image into a different coordinate space (thus affecting its physical meaning) - that is why we add both forward (HoughToRadon Transform after FHT) and transposed (RadonToHough Transform before TFHT) transform layers to prevent this. Table $1$ and Fig. \ref{ris:nn} show its layer-by-layer configuration. This construction allows the convolution layers $7-10$ to work with smaller feature maps that are more manageable than those after HT (as shown in Fig. \ref{ris:image}).

\medskip
\begin{center}
\scalebox{0.95}{
\begin{minipage}{\textwidth}
\begin{center}
\begin{tabular}[h]{| >{\centering}p{0.3cm}| >{\centering}p{1.5cm}| >{\centering}p{7.2cm}| >{\centering}p{3cm}|}
    \hline
    \multicolumn{1}{|c}{\textbf{№}} & \multicolumn{1}{|c}{\textbf{Layer type}} & \multicolumn{1}{|c}{\textbf{Parameters}} & \multicolumn{1}{|c|}{\textbf{Activation function}}\\
    \hline
    1 & conv & 4 filters 3×3, stride 1×1, padding 1×1 & softsign \tabularnewline
    2 & conv & 8 filters 3×3, stride 1×1, padding 1×1 & softsign \tabularnewline
    3 & conv & 16 filters 3×3, stride 1×1, padding 1×1 & softsign \tabularnewline
    4 & conv& 16 filters 3×3, stride 1×1, padding 1×1 & softsign \tabularnewline
    5 & FHT & - & - \tabularnewline
    6 & HRT & - & - \tabularnewline
    7 & conv& 16 filters 3×3, stride 1×1, padding 1×1 & softsign \tabularnewline
    8 & conv & 16 filters 3×3, stride 1×1, padding 1×1 & softsign \tabularnewline
    9 & conv & 16 filters 3×3, stride 1×1, padding 1×1 & softsign \tabularnewline
    10 & conv & 16 filters 3×3, stride 1×1, padding 1×1 & softsign \tabularnewline
    11 & RHT & - & - \tabularnewline
    12 & TFHT & - & - \tabularnewline
    13 & conv & 8 filters 3×3, stride 1×1, padding 1×1, upscale 2×2 & softsign \tabularnewline
    14 & conv & 4 filters 3×3, stride 1×1, padding 1×1, upscale 2×2 & softsign \tabularnewline
    15 & conv & 4 filters 3×3, stride 1×1, padding 1×1 & softsign \tabularnewline
    16 & conv & 2 filters 3×3, stride 1×1, padding 1×1 & softmax \tabularnewline
    \hline
  \end{tabular}

  \medskip
 Table 1. New architecture with HoughToRadon Transform.
   \end{center}
\end{minipage}}

\newpage
\begin{figure*}[h]
\centering
\scalebox{1.0}{\subfigure{
\includegraphics[height=0.35\textwidth]{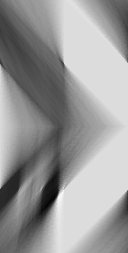}
\raisebox{0.0511\textwidth}{\includegraphics[height=0.2461\textwidth]{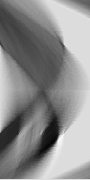}}
}
\subfigure{
\includegraphics[height=0.35\textwidth]{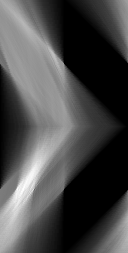}
\raisebox{0.0511\textwidth}{\includegraphics[height=0.2461\textwidth]{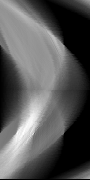}}
}}

\caption{Examples of 'inner'\ convolution feature maps from FHT (left) and FHT + HoughToRadon transform (right) layers at a consistent scale.}
\label{ris:image}
\end{figure*}
\end{center}

Our aim is to estimate the properties of this neural network, such as the error value and number of operations performed by the 'inner' convolution layers $7-10$, by modifying the input parameters $n$ and $scaleX$ of HoughToRadon transform layer. All neural networks were trained for 100 epochs under the same conditions, using the most common distortions such as adding noise, highlights, straight lines, blur and darkening. As a point of reference, we also compared new models with the default HoughEncoder model, which does
not include HoughToRadon transform.

\subsection{Dataset description}
For our experiments we use MIDV-$500$ \cite{lit17} dataset. It is an open dataset containing images of $50$ various types of documents taken at different angles with complex lighting and backgrounds, and the coordinates of the document corners. We will use the first $30$ of document types to train the neural network and the remaining $20$ to test. Before feeding images to the NN, we select only ones that have at least three corners of the document within the image, convert them to greyscale and scale all of them to $256 \times 256$ size. The total number of used images is $11965$, with the test part consisting of $4748$ images. 

\subsection{Evaluation metrics}
In order to compare the accuracy of the NN performance in a series of experiments, we use mean intersection over union (MIoU) distance. Its result is obtained using the equation \eqref{eq3}:

\begin{equation}\label{eq3}
MIoU = \frac{1}{N} \sum_{i=0}^{N-1} \frac{A_{i} \cap G_{i}}{A_{i} \cup G_{i}},
\end{equation}

where $N = 2$ (background and foreground), $A_{i}$ and $G_{i}$ are the output of the neural network and the actual answer for the relevant class, respectively.

\section{Results}\label{sec6}
Table $2$ summarizes the results of various runs. Each cell in the table contains three values: feature maps size, MIoU percentage and number of operations performed by the 'inner' convolution layers (up to multiplication by $10^{7}$).  The values higher than 97\% are emboldened.

To keep the experiment clean, we also compared the accuracy of the original HoughEncoder and the new neural network under the condition that sizes of the feature maps of the 'inner' convolutions coincide. To achieve that, we selected such $n$ and $scaleX$ coefficients so that sizes match. Results are presented in Table $3$. It is observed that coinciding in the intermediate feature maps sizes, HoughEncoder + HoughToRadon transform underperformed the original model while achieving state-of-the-art results with other combinations of input parameters.

\begin{table}[H]
\begin{center}
\scalebox{0.78}{\begin{tabular}{| >{\centering\arraybackslash}m{0.7in}| >{\centering\arraybackslash}p{0.6in}| >{\centering\arraybackslash}p{0.6in}| >{\centering\arraybackslash}p{0.6in}| >{\centering\arraybackslash}p{0.6in}| >{\centering\arraybackslash}p{0.6in}| >{\centering\arraybackslash}p{0.6in}| >{\centering\arraybackslash}p{0.6in}| >{\centering\arraybackslash}p{0.6in}| >{\centering\arraybackslash}p{0.6in}| >{\centering\arraybackslash}p{0.6in} |}
\hline
\diagbox[width=0.9in]{$n$}{$scaleX$} & 0.178 & 0.356 & 0.533 & 0.711 & 0.889 & 1.067 & 1.244 & 1.422 & 1.6 & 1.778\\
\hline
61 & [16; 61] \par 96.2\par0.2 & [32; 61] \par96.9 \par0.4 & [48; 61] \par\textbf{97.1} \par0.7  &  [64; 61] \par95.9 \par0.9 &  [80; 61] \par96.2 \par1.1 &  [96; 61] \par 95.0\par1.3  & [112; 61] \par95.8 \par1.6 & [128; 61] \par95.6 \par1.8 & [144; 61] \par96.0 \par2.0 & [160; 61] \par95.4 \par2.2\\
\hline
93 & [16; 93] \par\textbf{97.0} \par0.3 & [32; 93] \par \textbf{97.3} \par 0.7 & [48; 93] \par 95.3 \par 1.0 &  [64; 93] \par96.7 \par1.4 &  [80; 93] \par95.4 \par1.7 &  [96; 93] \par 95.6 \par 2.1 & [112; 93] \par95.2 \par2.4 & [128; 93] \par95.9 \par2.7 & [144; 93] \par95.3 \par3.1 & [160; 93] \par95.0 \par3.4\\
\hline
125 & [16; 125] \par96.2 \par0.5 & [32; 125] 96.7 \par 0.9 & [48; 125] \par 96.3 \par1.4 &  [64; 125] \par95.0 \par1.8 &  [80; 125] \par96.6 \par2.3 &  [96; 125] \par 96.4 \par2.8 & [112; 125] \par\textbf{97.1} \par3.2 & [128; 125] \par95.4 \par3.7 & [144; 125] \par94.8 \par4.1 & [160; 125] \par95.0 \par4.6\\
\hline
157 & [16; 157] \par96.9 \par0.6 & [32; 157] \textbf{97.2} \par 1.2 & [48; 157] \par96.1 \par1.7 &  [64; 157] 95.4 \par2.3 &  [80; 157] 96.2 \par2.9 &  [96; 157] 95.8 \par3.5 & [112; 157] 94.9\par 4.1 & [128; 157] 95.1\par 4.6& [144; 157] 94.9\par 5.2 & [160; 157] 95.1\par 5.8\\
\hline
189 & [16; 189] \par96.3 \par0.7 & [32; 189] 95.8 \par 1.4 & [48; 189] \par95.7 \par2.1 &  [64; 189] 96.0\par 2.8 &  [80; 189] 96.5\par 3.5 &  [96; 189] 96.7\par 4.2 & [112; 189] 95.2\par 4.9& [128; 189] 95.3\par 5.6& [144; 189] 94.4\par 6.3& [160; 189] 95.1\par 7.0\\
\hline
221 & [16; 221] \par95.6 \par0.8 & [32; 221] 96.2 \par 1.6 & [48; 221] \par96.2 \par2.4 &  [64; 221] 96.1\par 3.3 &  [80; 221] 96.7\par 4.1 &  [96; 221] 96.7\par 4.9 & [112; 221] 95.2\par 5.7& [128; 221] 95.2\par 6.5& [144; 221] 94.9\par 7.3 & [160; 221] 95.0\par 8.1\\
\hline
253 & [16; 253] \par95.9 \par0.9 & [32; 253] 95.9 \par 1.9 & [48; 253] \par\textbf{97.5} \par2.8 &  [64; 253] \textbf{97.7}\par 3.7 &  [80; 253] \textbf{97.0}\par 4.7 &  [96; 253] 96.8\par 5.6 & [112; 253] 95.5\par 6.5& [128; 253] 94.6\par 7.5& [144; 253] 95.0\par 8.4 & [160; 253] 94.8\par 9.3\\
\hline
285 & [16; 285] \par96.0 \par1.1 & [32; 285] 95.9 \par 2.1 & [48; 285] \par95.8 \par3.2 &  [64; 285] \textbf{97.0}\par 4.2 &  [80; 285] 96.2\par 5.3 &  [96; 285] 96.2\par 6.3 & [112; 285] 95.5\par 7.4& [128; 285] 95.1\par 8.4& [144; 285] 95.1\par 9.5& [160; 285] 94.5\par 11\\
\hline
317 & [16; 317] \par96.1 \par1.2 & [32; 317] 95.9 \par 2.3 & [48; 317] \par95.5 \par3.5 &  [64; 317] 96.7\par 4.7 &  [80; 317] 96.6\par 5.8 &  [96; 317] 96.2\par 7.0 & [112; 317] 94.9\par 8.2& [128; 317] 95.1\par 9.3& [144; 317] 94.5\par 11& [160; 317] 94.9\par 12\\
\hline
349 & [16; 349] \par96.2 \par1.3 & [32; 349] 96.0 \par 2.6 & [48; 349] \par96.9 \par3.9 &  [64; 349] \textbf{97.6}\par 5.1 &  [80; 349] 96.5\par 6.4 &  [96; 349] 96.4\par 7.7 & [112; 349] 96.0\par 9.0& [128; 349] 96.3\par 10& [144; 349] \textbf{97.5}\par 12& [160; 349] 96.9\par 13\\
\hline
\end{tabular}}
\medskip
\caption*{Table 2. Experimental results.}
\end{center}
\end{table}

\begin{center}
\begin{figure*}[h]
\centering
\scalebox{0.8}{
\subfigure{
\includegraphics[height=0.25\textwidth]{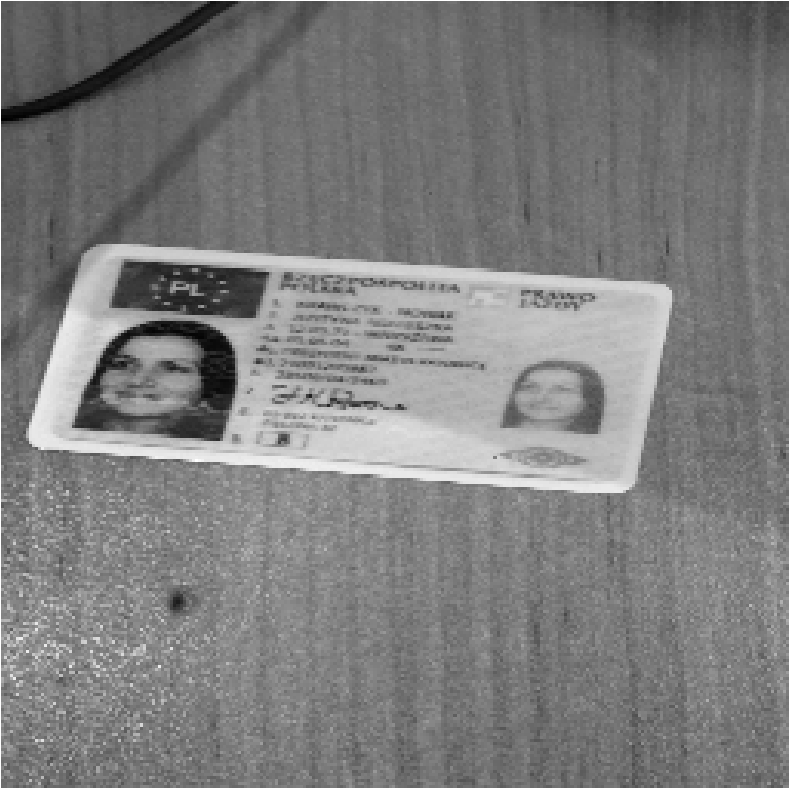}
\includegraphics[height=0.25\textwidth]{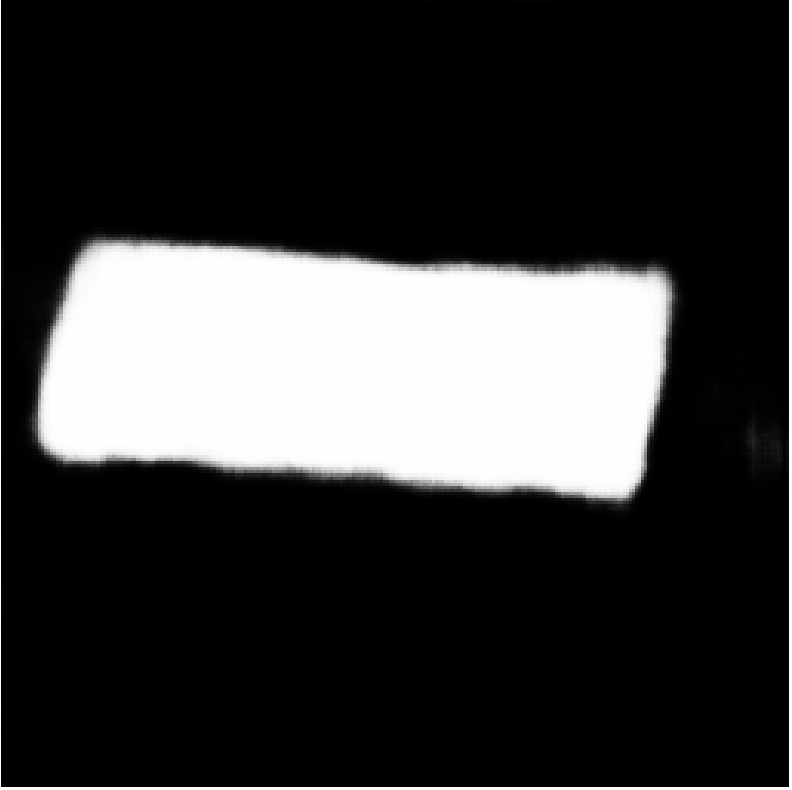}
\includegraphics[height=0.25\textwidth]{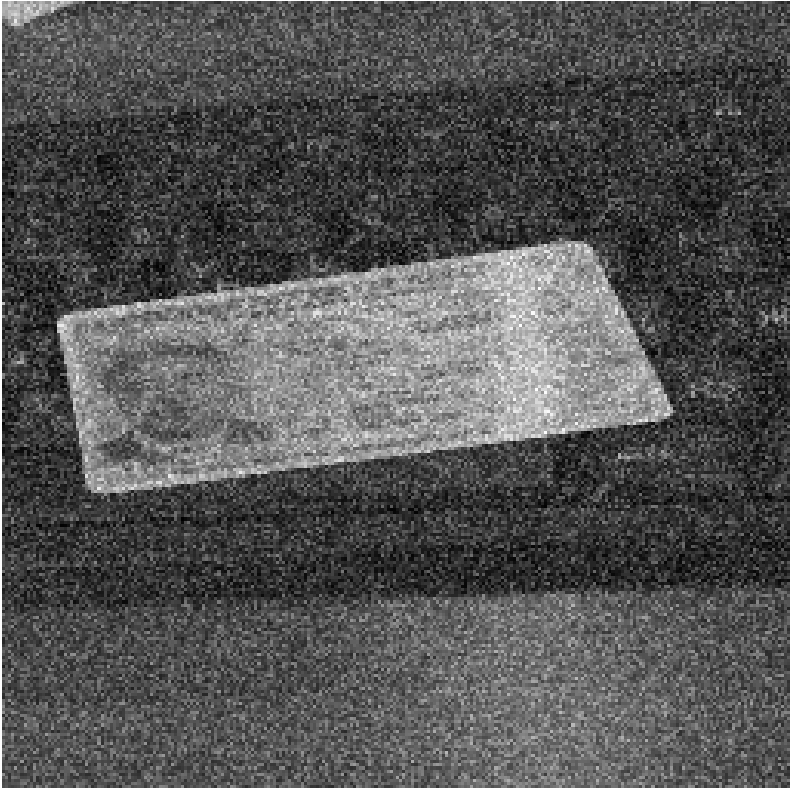}
\includegraphics[height=0.25\textwidth]{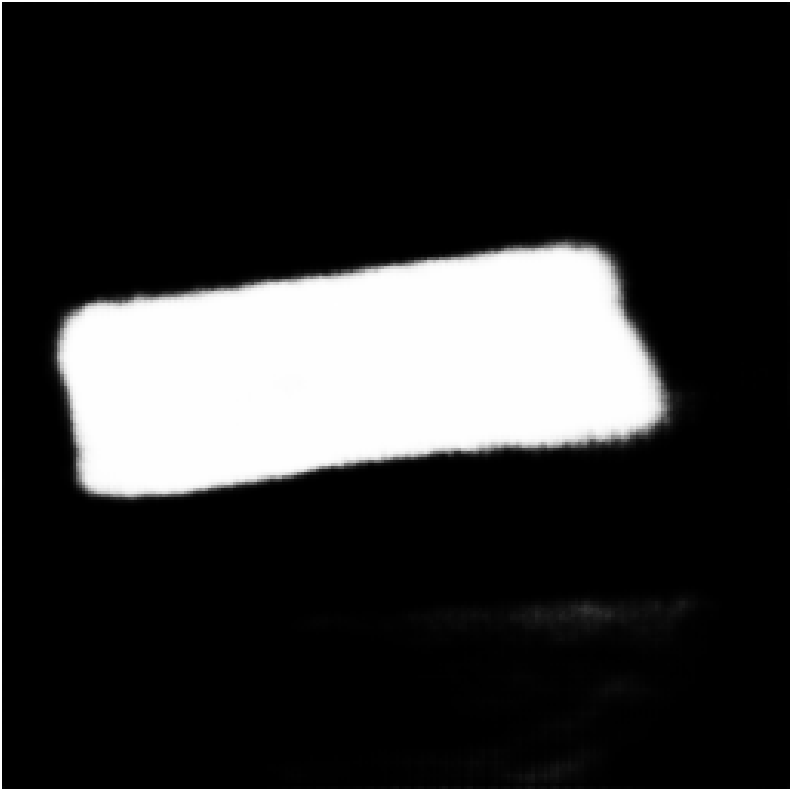}
}}
\scalebox{0.8}{
\subfigure{
\includegraphics[height=0.25\textwidth]{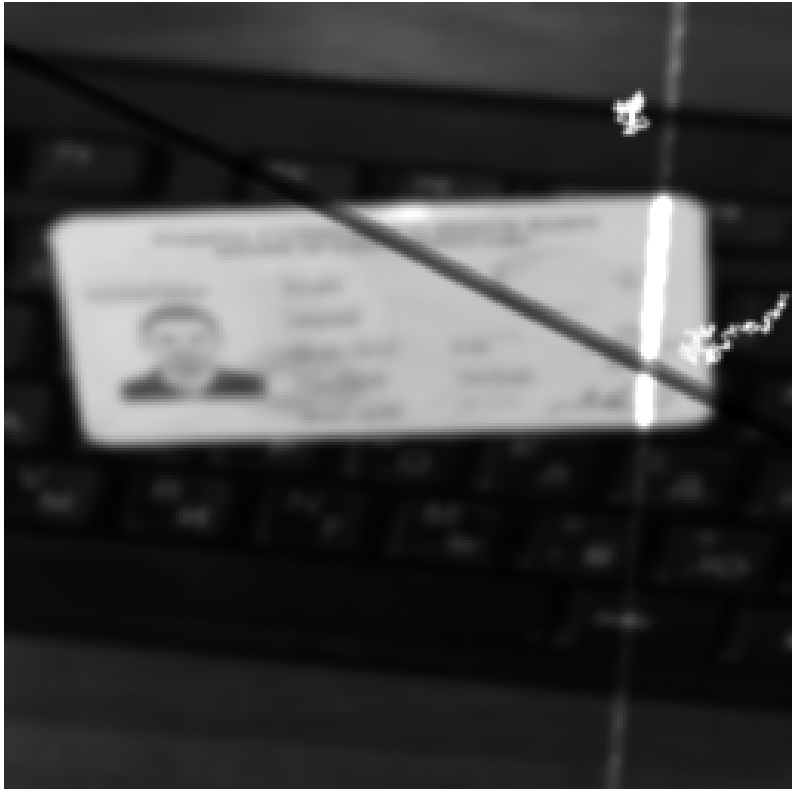}
\includegraphics[height=0.25\textwidth]{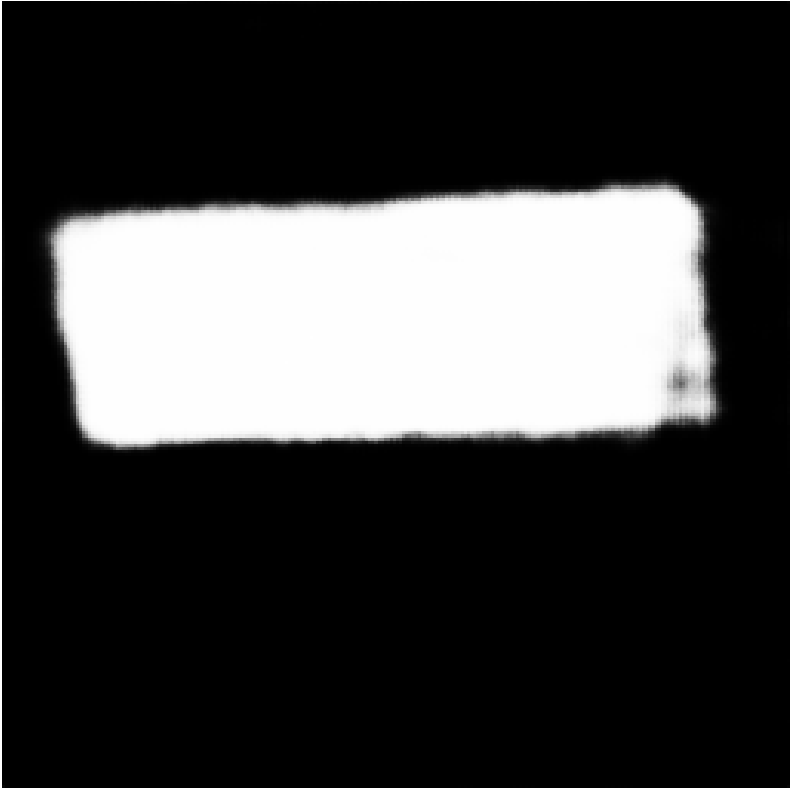}
\includegraphics[height=0.25\textwidth]{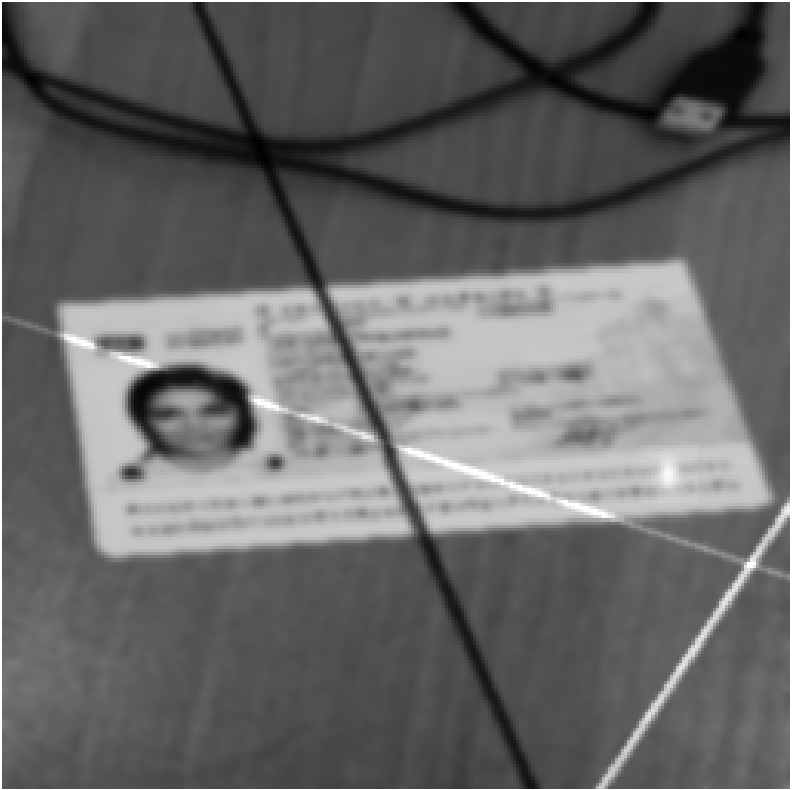}
\includegraphics[height=0.25\textwidth]{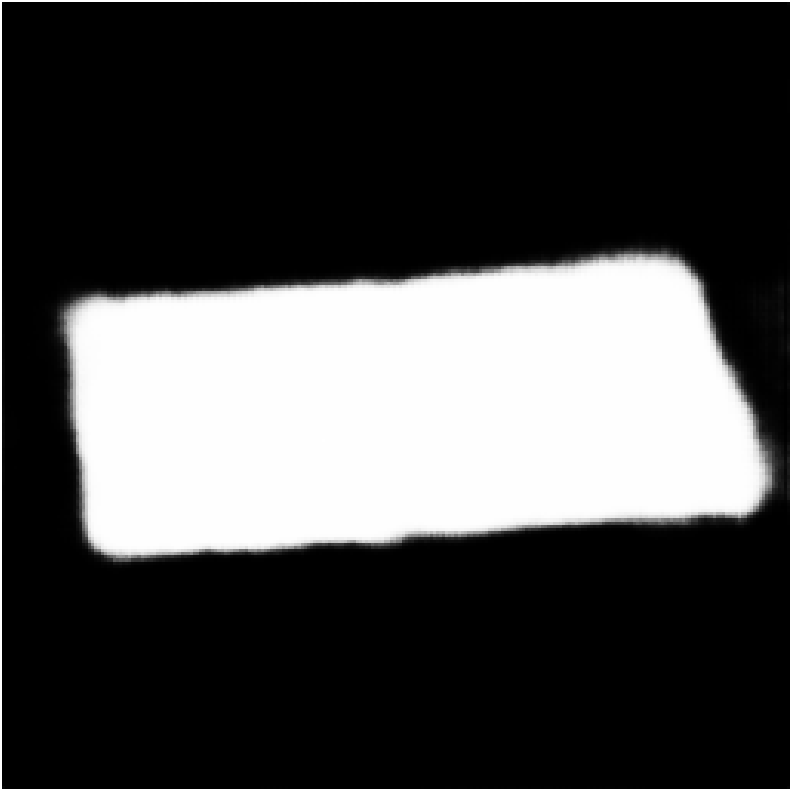}
}}

\caption{Examples of input and output images of the NN with HoughToRadon transform.}
\label{ris:image3}
\end{figure*}
\end{center}

The results presented in Table $2$ highlight that HRT and RHT layers serve their purpose and allow the neural network to solve the task while doing much fewer operations. With the use of $n$ and $scaleX$ parameters, we can vary the size within a wide range of values to optimize computational complexity. The comparison of the MIoU values proves that decreasing the sizes of intermediate feature maps does not always result in a loss of quality. It can be seen that we can obtain around $97\%$ savings in number of operations with even $0.2\%$ gain in accuracy compared to HoughEncoder. The NN can provide the same quality with different numbers of operations, so there is no need to perform more operations to solve the task. Furthermore, it is clear from the results that our approach not only saves time and reduces memory requirements, but also boosts the accuracy of segmentation from $96\%$ \cite{lit10} up to $97.7\%$. Fig. \ref{ris:image3} shows some results of semantic document segmentation generated by the NN featuring HoughToRadon transform.

\medskip
\begin{center}
\begin{minipage}{0.75\textwidth}
\begin{center}
\begin{tabular}{| >{\centering\arraybackslash}p{1.2cm}| >{\centering\arraybackslash}p{1.2cm}| >{\centering\arraybackslash}p{2.5cm}| >{\centering\arraybackslash}p{2.0cm}| >{\centering\arraybackslash}p{3.0cm}| }
    \hline
    \multicolumn{1}{|c}{\begin{tabular} {@{}c@{}} n \end{tabular}} & \multicolumn{1}{|c}{\begin{tabular} {@{}c@{}} scaleX \end{tabular}} & \multicolumn{1}{|c}{\begin{tabular} {@{}c@{}} Feature maps size \end{tabular}} & \multicolumn{1}{|c}{\begin{tabular} {@{}c@{}} MIoU \end{tabular}} & \multicolumn{1}{|c|}{\begin{tabular} {@{}c@{}} Operations / $10^{7}$ \end{tabular}}\\
    \hline
    \multicolumn{5}{|c|}{HoughEncoder}\\
    \hline
    - & - & [128; 253] & 96.0 & 7.5\\
    \hline
    \multicolumn{5}{|c|}{HoughEncoder + HoughToRadon}\\
    \hline
    253 & 1.422 & [128; 253] & 94.6 & 7.5\\
    \hline
  \end{tabular}
  
\medskip
 Table 3. Comparison between HoughEncoder and HoughEncoder + HoughToRadon, matching feature maps sizes.
   \end{center}
\end{minipage}
\end{center}

\newpage
\section{Conclusion and future work}\label{sec7}

In this paper, we have introduced a new neural network layer named HoughToRadon transform. It is responsible for transforming the image coordinates from ($s, t$) to ($\rho, \varphi$) space and back. Layer's parameters $n$ and $scaleX$ allow convolutional filters to use much smaller feature maps, thereby shortening the time needed to train the neural network. By integrating it into the HoughEncoder architecture \cite{lit10}, we have not only achieved a significant reduction in number of operations, but have also improved the quality of document detection, which proves the effectiveness of the newly introduced layer.

In our following research, we intend to switch to more complex tasks of semantic image segmentation and find cases where our approach leads to quality compromising. Also, we plan to determine whether the acquired linearity of the new coordinate space by angle and shift affects the detection accuracy.

	\bibliographystyle{spiebib}

\begin{thebibliography}{}
    \bibitem{lit1}  Li, Q., Cai, W., Wang, X., Zhou, Y., Feng, D.D., and Chen, M. (2014) “Medical image classification with convolutional neural network”, 2014 13th International Conference on Control Automation Robotics \& Vision (ICARCV), Singapore. doi: 10.1109/ICARCV.2014.7064414
    \bibitem{lit2} Jiang Y, Chen L, Zhang H and Xiao X (2019) “Breast cancer histopathological image classification using convolutional neural networks with small SE-ResNet module”, PLoS ONE 14(3): e0214587. doi: 10.1371/journal.pone.0214587
    \bibitem{lit3} Bruno D.R., Osorio F.S. (2017) “Image classification system based on deep learning applied to the recognition of traffic signs for intelligent robotic vehicle navigation purposes”, Latin American Robotics Symposium (LARS) and 2017 Brazilian Symposium on Robotics (SBR), Curitiba, Brazil. doi: 10.1109/SBR-LARS-R.2017.8215287.
    \bibitem{lit4} Madan R., Agrawal D., Kowshik S., Maheshwari H., Agarwal S., and Chakravarty D. (2019) “Traffic Sign Classification using Hybrid HOG-SURF Features and Convolutional Neural Networks”, in ICPRAM. doi: 10.5220/0007392506130620
    \bibitem{lit5} Lecun Y., Bottou L., Bengio Y. and Haffner P.  (1998) “Gradient-based learning applied to document recognition”, in Proceedings of the IEEE, doi: 10.1109/5.726791
    \bibitem{lit6} Jnawali K., Arbabshirani M.R., Rao N., and Patel A.A. (2018) “Deep 3D convolution neural network for CT brain hemorrhage classification”, Proc. SPIE 10575, Medical Imaging 2018: Computer-Aided Diagnosis; doi: 10.1117/12.2293725
    \bibitem{lit7} Sheshkus A., Ingacheva A., and Nikolaev D. (2018) “Vanishing points detection using combination of fast Hough transform and deep learning”, Proc. SPIE 10696, Tenth International Conference on Machine Vision (ICMV 2017); doi: 10.1117/12.2310170
    \bibitem{lit8} Prun V.E., Nikolaev D.P., Buzmakov A.V. et al. (2013) “Effective regularized algebraic reconstruction technique for computed tomography”, Crystallogr. doi: 10.1134/S1063774513070158
    \bibitem{lit9} Karpenko S., Nikolaev D. (2008) “Hough Transform: Underestimated Tool In The Computer Vision Field”, 22nd European Conference on Modelling and Simulation, ECMS 2008. doi: 10.7148/2008-0238
    \bibitem{lit10} Sheshkus A., Nikolaev D., and Arlazarov V. (2020) “Houghencoder: Neural Network Architecture for Document Image Semantic Segmentation”, 2020 IEEE International Conference on Image Processing (ICIP). doi: 10.1109/ICIP40778.2020.9191182
    \bibitem{lit11} Aliev M., Ershov E.I., and Nikolaev D.P. (2018) “On the use of FHT, its modification for practical applications and the structure of Hough image”, International Conference on Machine Vision. https://arxiv.org/pdf/1811.06378.pdf
    \bibitem{lit12} M. A. Aliev, D. P. Nikolaev, A. A. Saraev (2014) "Construction of fast computing adjustment for Niblack binarization algorithm",  ISA RAN V. 64.3. P. 25–34.
    \bibitem{lit13} Ballard D.H. (1981) “Generalizing the Hough transform to detect arbitrary shapes”, Pattern Recognition, 13.2, ISSN 0031-3203. doi: 10.1016/0031-3203(81)90009-1
    \bibitem{lit14} Ershov E.I. (2017) “Generation Algorithms Of Fast Generalized Hough Transform”, 31st European Conference on Modelling and Simulation, ISBN: 978-0-9932440-4-9
    \bibitem{lit15} Sheshkus A.V., Ingacheva A., Arlazarov V.L., and Nikolaev D.P. (2019) “HoughNet: Neural Network Architecture for Vanishing Points Detection”, 2019 International Conference on Document Analysis and Recognition (ICDAR).  doi: 10.1109/ICDAR.2019.00140
    \bibitem{lit16}  Dolmatova A. V., Nikolaev D. P. "Uskorenie svertki i obratnogo proetsirovaniya pri rekonstruktsii tomograficheskikh izobrazhenii" [Fast filtering and back projection for ct image reconstruction]. Sensornye sistemy [Sensory systems]. 2020. V. 34(1). P. 64–71 (in Russian). doi: 10.31857/S0235009220010072
    \bibitem{lit17} Arlazarov V.V., Bulatov K.B., and Chernov T.S. (2018) "MIDV-500: A Dataset for Identity Documents Analysis and Recognition on Mobile Devices in Video Stream". ArXiv: 1807.05786.
    \bibitem{lit18} ShylajaS. S., Murthy K.N., and Natarajan, S. (2011) "Feed Forward Neural Network Based Eye Localization and Recognition Using Hough Transform”, International Journal of Advanced Computer Science and Applications, 2. doi: 10.14569/IJACSA.2011.020318
    \bibitem{lit19} Wong A., and Bishop W. (2008) “Robust Hough-Based Symbol Recognition Using Knowledge-Based Hierarchical Neural Networks”, International Conference on Image Processing, Computer Vision, \& Pattern Recognition. ISBN: 1-60132-078-7
    \bibitem{lit20} Bailey D.G., Chang Y., and Moan S.L. (2020) “Analysing Arbitrary Curves from the Line Hough Transform”, Journal of Imaging, 6. doi: 10.3390/jimaging6040026
    \bibitem{lit21} Goodfellow I., Bengio Y., and Courville A. (2016) "Deep Learning", MIT Press, url: http://www.deeplearningbook.org
    \bibitem{lit22} Bichen Wu, Chenfeng Xu, Xiaoliang Dai, Alvin Wan, Peizhao Zhang, Masayoshi Tomizuk, Kurt Keutzer, Peter Vajda. (2020) "Visual Transformers: Token-based Image Representation and Processing for Computer Vision", Facebook AI, UC Berkeley. ArXiv: 2006.03677v4

\end{thebibliography}

\end{document}